# It is not "accuracy vs. explainability" – we need both for trustworthy AI systems

Prof. D. Petkovic

CS Department
San Francisco State University
USA
petkovic@sfsu.edu

## Abstract

We are witnessing the emergence of an "AI economy and society" where AI technologies are increasingly impacting health care, business, transportation and many aspects of everyday life. Many successes have been reported where AI systems even surpassed the accuracy of human experts. However, AI systems may produce errors, can exhibit bias, may be sensitive to noise in the data, and often lack technical and judicial transparency resulting in reduction in trust and challenges in their adoption. These recent shortcomings and concerns have been documented in scientific but also in general press such as accidents with self-driving cars, biases in healthcare, hiring and face recognition systems for people of color, seemingly correct medical decisions later found to be made due to wrong reasons etc. This resulted in emergence of many government and regulatory initiatives requiring *trustworthy and ethical AI* to provide *accuracy and robustness, some form of explainability, human control and oversight, elimination of bias, judicial transparency and safety*. The challenges in delivery of trustworthy AI systems motivated intense research on explainable AI systems (XAI). Aim of XAI is to provide human understandable information of how AI systems make their decisions. In this paper we first briefly summarize current XAI work and then challenge the recent arguments of "accuracy vs. explainability" for being mutually exclusive and being focused only on deep learning. We then present our recommendations for the use of XAI in full lifecycle of high stakes trustworthy AI systems delivery, *e.g. development; validation/certification; and trustworthy production and maintenance.* While most of our examples and discussion relate to health and biomedical application due to their high impact to society and economy and large body of published work, we believe that our recommendations are relevant to delivery of all types of high stakes trustworthy AI applications.

## 1. Introduction and background

We are witnessing the emergence of an "AI economy and society" where AI technologies are increasingly impacting health care, business, transportation and many aspects of everyday life. Many successes have been reported where AI systems even surpassed the accuracy of human experts. However, it has been discovered that AI systems may produce errors, can exhibit bias, may be sensitive to noise in the data, and often lack technical and judicial transparency resulting in reduction in trust and challenges in their adoption.  These recent shortcomings and concerns have been documented in scientific but also in general press such as accidents with self-driving cars, biases in healthcare, hiring and face recognition systems for people of color, seemingly correct medical decisions later found to be made due to wrong reasons etc. (1, 2, 3, 4, 5 ,6, 15, 16). This resulted in emergence of many government and regulatory initiatives to ensure that high risk and impact AI applications become *trustworthy and ethical* by providing the following common components: *accuracy and robustness, transparency and explainability, human control and oversight, fairness and elimination of bias, and mitigation of risk and safety*  (7, 8, 9). Some like EU GDPR (9) already became a law and are now expanded into proposed EU AI Act legislation (7), expecting to influence similar legislations in the rest of the world. More concrete best practices based on these regulations are being developed as well (10, 11, 12, 13, 36). Legal considerations related to liability, contract and tort law are also being discussed and may settle some open issues via courts (14).

Explainability and transparency are figuring in some form and level in all of the above mentioned regulatory initiatives for trustworthy AI, proportional to estimated level of risk and safety issues of related applications. This motivated intense research on explainability of AI systems (XAI) (17, 18, 19, 23, 25, 36). Aim of XAI is to provide human understandable information of how AI systems make their decisions, a challenging task. XAI task is further complicated due to intricate interaction between AI systems and human users both in terms of decision making as well as in terms of evaluating and understanding the explanations themselves, as analyzed in (24, 34, 36).The very need for XAI  is not without some controversy: while many researchers agree with the need for XAI especially for high stakes AI applications, number of researchers advocate the use of black box (mainly deep learning AI) approaches even if they are not explainable, arguing that this is justified by their higher accuracy and that insisting on XAI actually impedes the benefits and potential of AI (21, 22). On the other hand, some researchers even advocate abandonment of black box systems altogether and using only explainable ones (23), a difficult proposition given plethora of well-developed and studied AI algorithms. The tension on whether we need XAI or not is often framed as "accuracy *vs.* explainability", in other words accuracy and explainability are discussed as mutually exclusive. This discussion most often involves AI deep learning applications in healthcare and biomedicine due to their high  impact to people and economy and significant risk and safety issues.

In this paper we first briefly summarize current XAI work and then challenge the recent arguments of "accuracy vs. explainability" for being mutually exclusive and being focused only on deep learning. We then present our recommendations for integration of XAI analysis (e.g. application of various XAI methods) in all lifecycle components of high stakes trustworthy AI systems:  *development; validation/certification; and trustworthy production and maintenance.* As in (7), instead of AI algorithms, we focus on *AI systems*, defined as complex systems of "feature capture" devices, coupled with AI algorithms running on supporting software and hardware. While most of our examples and discussion relate to health and biomedical application due to their high impact, risk and safety issues as well as large body of published work, we believe that our recommendations are relevant to delivery of all types of high stakes trustworthy AI applications.

## 2. Brief summary of XAI advances today

AI systems at high level can be divided into *feature-based systems* where the input data is mostly in tabular form with 10s or 100s of distinct features (measures of some variable like medical tests) which are either pre-existing or designed by humans, and now popular *deep learning systems* (usually in the form of convolutional neural networks) where inputs are images and signals with millions of data elements or pixels which are fed directly to AI systems, with no human involvement in feature design. One benefit of deep learning systems is that there is no need for human intensive "feature engineering", since they can discover features not discoverable by humans. These features, present in internal layers of deep neural networks are however hard to observe and understand (in contrast to feature based system), and this in turn hampers their explainability. XAI techniques then aim to explain how AI systems make their decisions, and aim to do so in easily understandable way to humans (17, 18, 19, 23, 25, 26, 27, 36). XAI in general does not aim to provide any causality or interpretability analysis (e.g. reasons *why* decisions have been made) but simply aims to provide factual information about specific ways AI algorithms make decisions on given data. XAI methods generally involve the analysis of already trained AI models and they can be *global or model* based (explaining how AI model a*s a whole* works on for example totality of training data) or *sample* based *or local* (how AI classified a specific sample). They can be *algorithm agnostic* (independent of specific AI algorithm) or *algorithms specific*. Usually, agnostic XAI algorithms (LIME being most popular (25)) are based on some form of local approximation or simulation of AI algorithm they try to explain, an issue that may cause problems (23). Algorithm specific explainers leverage the trained AI system they aim to explain (e.g. 26). Most of XAI methods identify and rank "most predictive features" or image regions ("salient image regions") by using type of a sensitivity analysis e.g. perturbing the data and observing changes in accuracy, where those features causing larger change in accuracy are ranked higher. In general, achieving satisfactory XAI is much more challenging for deep learning systems – majority of deep learning XAI methods are limited to *sample or local explainers* and most of them only point to what image regions (pixels) have been used for classification of a sample and not how the classification has actually been done. Feature based systems often offer more XAI information, especially for tree based systems (e.g. 26). XAI is relatively new field and intense research efforts in XAI especially for deep learning systems are ongoing (17, 18, 19). One type of XAI information often forgotten in academic works which we deem critical for risk and safety mitigation of trustworthy AI systems is *classification confidence*, true and tried method in other analytical disciplines. Issues of human interaction with AI systems (humans being evaluators, auditors or end users) in decision making as well as in evaluating usefulness of XAI information is very complex as pointed out in (36), and is beyond the scope of this paper, except in pointing out to importance of considering human users in all aspects of XAI development and usage.

## 3. "Why sacrifice accuracy because of explainability"- arguments for and against

We now summarize and challenge the key arguments of proponents of "no need for XAI for highly accurate (but hard to explain) deep learning black box systems".

*Humans trust decisions even if they do not know how they are made*: This argument states that insisting on XAI has its perils and that as long as non-explainable black box systems are accurate and well tested they should be used (the issue of "well tested" is critical but not trivial and will be addressed below) (21, 22). We first note that this analogy is being made wrt. human reaction to decisions made by humans (e.g. via advice of experts leveraging historical experience) which is not completely proven to hold for decisions made by AI systems. Humans may be more concerned with non-explainable decisions if they are made by some AI system, and especially in the presence of mistakes made on the data/entities they understand (15, 24, 25, 36). This is evidenced by strong public backlash related to documented errors and bias of AI systems in for example face recognition, hiring, self-driving car accidents etc. (4, 5, 15, 16). Importantly however, we note that most human users do trust machines which they do not understand well as long as they are certified by a trusted agency (for example how many people know how autopilots work but they still use air travel, knowing that planes are certified). This then brings us back to the need for some form of independent validation or certification of high stakes AI systems as proposed by most approaches for achieving trustworthy AI.

*Deep learning systems are highly accurate and robust:* There are several issues of concern here. We first observe that some AI deep learning applications that seemed to have achieved high accuracy were found later, using some form of XAI analysis, to have done it "due to all the wrong reasons" (1, 3, 6). In one example from (1), deep learning medical imaging applications achieved seemingly very accurate decisions by using information related to X-ray machine type and not patient information (reason was that sick and healthy patients were imaged using consistently different X-ray machines). Notably as well, deep leaning systems are expensive to train and require huge amount of quality training data and often require help of special hardware. The discussion on how much training data we need for proper AI model training is complex and depends on AI model complexity, number of decision classes and statistics of the data (good discussion is in (32)). From complex theoretical analysis, researchers developed some empirical rules on minimal size of training data, two most often mentioned being as follows: a) we need at minimum order of magnitude (10X) more training data than degrees of freedom of AI model (e.g. number of trainable parameters); b) we need minimum of 1000 training samples per decision class. Given that most feature based AI models have 10-100s of features while deep leaning has often 10s of millions of trainable neural connections, this points to the need for huge training data sets in the latter case, often 5-6 orders of magnitude larger than for feature based systems. Training AI models of high complexity or "degrees of freedom" with insufficient data causes problems in reliable accuracy estimation like *overfitting* and is often called "curse of dimensionality" (33). As a result, deep learning systems often lack robustness (29), which in turn may open them to adversarial attacks (31) – one more reason for extensive training and testing. Hence, achieving true high accuracy and robustness of deep learning systems for future (unseen) and noisy data is a nontrivial challenge and is generally much more difficult than for feature based systems.

*It is all about deep learning.* Focus on popular deep learning approaches which are best applied to image based AI applications such as face recognition, medical image diagnostic etc. is missing many important and very accurate feature (tabular) AI algorithms and applications (many of them more explainable), as shown in review of 13 AI methods on 165 classification problems in biomedical area (20).

## 4. Role of XAI analysis in delivery of trustworthy and cost effective AI systems

We now outline critical role XAI analysis can play in delivery of trustworthy and cost effective AI systems of any type (deep learning or feature based) in all of their lifecycle components: a) *development* including internal testing; b) independent external *verification, audit or certification* (in our opinion inevitable for high stakes AI applications)*; and finally in c) trustworthy production and maintenance.* XAI analysis is therefore proposed as a necessary step to be integrated with current best practices recommended for all types of high stakes AI systems. We relate our recommendations to key components of trustworthiness for AI systems mentioned above: *accuracy and robustness, transparency and explainability, human control and oversight, fairness and elimination of bias, and mitigation of risk and safety*.

**4.1 XAI analysis in development and internal testing stage**

*Exploring tradeoffs between AI model complexity and loss of accuracy:* In this fundamental step, attempts should be first made to reduce complexity of trained AI models by leveraging XAI to focus on reduced number of most predictive features or image salient regions and understand how these relate to achievable (e.g. loss) of accuracy. For example, we observed, based on our work on several biomedical applications of Explainable Random Forest (RFEX) on tabular (feature) data, that only 2-4 percent of top ranked (most predictive) features achieved almost the same classification accuracy as by using all features, allowing us to establish some form of *tradeoff function between subsets of features (best 2, best 3 etc.) and achievable accuracy*, called "cumulative F1 score" (26, 27). This allows for example choosing minimal complexity model for desired accuracy. While harder, this approach can also be applied to deep learning AI applications, by focusing on reduced number of highly predictive salient mage regions, for example by using preprocessing to first find image regions containing heads/faces or important image regions. It is not only that simpler AI models are easier to test and explain, but also that they provide more robust accuracy estimates given larger ratio of the number of training samples to number of features of "degrees of freedom". This in turn mitigates serious consequences of "curse of dimensionality" as noted above. (Occam's razor paradigm also comes to mind (28), stating that simpler explanations of data are more reliable). This initial reduction of model complexity wrt. accuracy has impact on many subsequent stages in AI systems lifecycle.

*Elimination of noisy training data or "outliers":* Availability of quality training data with representative features/inputs and correct ground truth labels is critical for successful training and optimization of AI systems. As said before, AI systems are complex senor/hardware/software systems and nosy or corrupted data can often happen. Detecting possible "outliers" (erroneous feature values or ground truth) is a common practice today before AI training phase, but it is challenging. Sample based XAI analysis helps here by for example identifying samples classified with low confidence and/or sample features values inconsistent with explanation models (one such attempt is done in (26)). This analysis can also point to wrong ground truth labels (often the case when ground truth is determined for health applications by human/expert voting).

*Accuracy statistics are not enough, we need to check for consistency with domain knowledge*: We recommend that XAI analysis be used as a mandatory checking/validation tool in addition to standard measures of AI systems accuracy (e.g. precision, recall, F1 score, sensitivity, specificity, ROC, AUC) to ensure that most predictive features or image salient regions are indeed consistent with application domain specifcs or conversely checking if they are obviously not consistent. While important for all types of AI systems, this validation is especially important for deep learning systems since they use features which are hard to verify and understand. XAI can for example answer simple but important questions helpful for increasing the trust of domain or certification experts: a) "Did AI model (or its simpler explainable version) base its decisions on features or image regions consistent with application domain", or point to the opposite: b) "The system made its decisions on obviously wrong or irrelevant features or image regions" (e.g. patient ID, hospital name, image background and not the object of interest etc.). While it is possible that deep learning can discover new features by itself, the latter type of check (as b) above) ensures elimination of obviously erroneous reasons for proper classification, which was shown to happen (1,3, 6). Good example of using XAI in addition to classic accuracy analysis to build user confidence for deep learning imaging application for detecting skin cancer is in (30), where in addition to classic accuracy measures (accuracy, ROC curves etc.) it is also shown that deep learning indeed used the appropriate image regions for its analysis (e.g. pixels related to suspected skin cancer sites under investigation). For feature based AI systems, XAI analysis can also help establish consistency between highly predictive top ranked features and ground truth established independently, as in (26), or prove that classification models made seemingly correct decisions but based on wrong features (25).

*Bias detection and suppression:* All approaches to achieve this are based on some form of XAI analysis and include d*isparate impact testing (*for example suppressing or removing features related to discriminatory and legally protected categories e.g. sex, age, sexual orientation and observing related changes in classification), or by observing XAI provided ranking of variables/features or salient image regions and checking if they are related to discriminatory categories (18). Given the future where AI systems will likely be audited, certified or challenged (7, 14), it is hard to imagine non-explainable AI applications being able to withstand these challenges. This step in turn may also help in auditing of training data for inherent bias (training data is considered as one of the biggest reasons behind the bias in AI systems (2) ).

*Enabling right to know, and judicial transparency:* During development stage it is essential to ensure that the above properties can indeed be provided by AI system being developed as they are required by most mentioned guidelines for trustworthy AI and envisioned legal challenges (7-14). XAI sample based methods are of great importance here due to their ability to offer insights into how specific decisions (e.g. those challenged by affected users) have been made.

*Risk, human oversight and safety:* At this stage assessment has to be made of the impact and limitations of AI system due to its decision errors under normal conditions, but also due to noisy or missing features and HW/SW errors. This is critical in high stakes applications like health care, robotics, self-driving cars and features prominently in all trustworthy AI regulatory initiatives. For example, EU AI Act (7) is driven by risk level of AI systems. XAI information (e.g. simplified AI models, ranked prediction variables or salient image regions, and *importantly* classification confidence levels) is critical by helping developers focus on most important components of the whole AI system (e.g. most predictive features or image regions) when analyzing the impact of noisy features or signals to the AI system. Similarly, XAI provided confidence level for sample classification can be used to prevent risky and unsafe decisions or actions as clearly required in (7).

XAI tools and information at this stage have to be designed, evaluated and documented in the form easily understandable to envisioned users such as auditors and end-users who may not necessarily be AI experts (as pointed out in 7, 36).

**4.2 XAI analysis in external verification, audit and certification stage**

In this stage, role of XAI analysis is similar as in the stage above, but is performed by external certification or audit agencies and domain experts on validation data not used in the above internal development stage, as per guidance such as in (7-14, 36) and proportional to anticipated risk and safety issues. This stage should leverage XAI tools from previous stage, as well as other independently provided XAI tools at the discretion of certification and audit organization.

*Validation of accuracy, auditability: A*ll regulations and best practices include the need for some form of explainability or transparency as integral part of establishing user trust and esnuruing ability to audit AI systems. At this stage, users - domain experts, or auditors' view of AI system accuracy on samples they know about (e.g. validation data sets) may well be a determining factor in forming their trust (23, 24, 25, 36).  As in the development stage, XAI analysis can provide critical information to check for consistency or obvious inconsistency in ways AI systems make their decisions.

*Audits for bias, right to know, human oversight, risk and safety.* In this stage similar XAI analysis reported in development phase can be applied here, but using independently provided verification data with feedback from auditors and domain experts who may not be experts in AI.

## 4.3 XAI analysis in production and maintenance phase

While not often addressed in academic research, this phase is critical for commercial success of AI and it benefits from XAI. Production AI systems should leverage XAI tools and information as provided from development stage. Notably, proper risk and safety are key concerns in production stage, and the level of XAI requirements is by many recommended to be proportional to estimated level of risk and safety exposures of specific AI application (7, 36).

*Robustness, human oversight, risk and safety:* As said before, AI systems are complex engineering products with many possibilities of errors or failures. Once we know relatively small number of most important features or image regions from XAI analysis, we can improve its robustness and mitigate risk and safety issues by ensuring their reliable and robust capture and processing. In addition, use of XAI provided confidence levels is now critically important to prevent possibly wrong actions or decisions due to unreliable decision making of AI system, an important ability to "pull the plug" as quoted in (7) as one of the ways to mitigate risk and safety.

*Right to know, auditability, bias and fairness, judicial transparency:* These issues may arise from legal or similar issues requiring formal action resulting from mistakes or errors AI production system made. It is obvious that related legal actions will require some form of sample based XAI analysis of disputed decisions as reported in (14). For example, legal investigations related to self driving car accidents are already happening and involve XAI analysis (e.g. (16) ).

*User interfaces for end users*: While not always required (e.g. depending on importance and risk impact of the AI application), XAI type of information for end user is believed to have a potential to improve their trust and decision making (17, 23, 25, 26, 36). As said before, interaction between end-users and AI systems is complex (36) and sometimes XAI information may produce "anchoring effect" e.g. decision biases (34). While we believe XAI to be promising here, more work needs to be done to investigate and evaluate utility of XAI for end users and specific applications, as well as to employ tried methods of User Centered Design (35) to adapt and optimize XAI outputs to the needs of end users, as attempted for example in (26, 27) and analyzed in more details in (36).

*Cost effective delivery:* Cost and hardware requirements related to extracting and storing large number of features, and associated complexity of AI run time engine are important for applications where AI has to be delivered efficiently such as to mobile devices. This can be helped again by XAI using tradeoffs between features or system complexity and achievable accuracy to choose optimal cost/benefit configuration of AI systems. Think how effective the deployment can be if one can deliver for example 98% of full accuracy with only few (vs. 100s) of features/sensors, with related reduction in AI algorithm memory and CPU requirements.

*Post-launch maintenance:* XAI analysis is also necessary in "post deployment" for dealing with AI system maintenance in case of errors in production (e.g. 7, 13) - understanding why samples were misclassified and using that to improve deployed AI system. Most of such tasks include applying sample explainers to samples of interest (e.g. samples with disputed decisions etc.).

Finally, we also mention two challenges reported in (23) where XAI could jeopardize commercialization of AI systems. One is the question arising from the fact that if explanations about a commercial AI system decision making are known, what would prevent others to copy it and provide competitive solutions. In this case AI industry may be de-incentivized to provide XAI information or invest in AI development. The second challenge is that XAI info could be used to defeat AI systems and fool them into making decisions benefiting malicious actors, although work has been reported that adversarial AI methods can also be used to help XAI (31). These are open challenges but we believe that they can be solved in a way similar to how analogous problems have been solved so far: for example XAI info can be kept confidential for auditing agencies only, malicious actors can be persecuted, and patent law can be used to protect AI designs.

## Acknowledgements

We are grateful to Dr. Ljubomir Buturovic for valuable feedback and to Prof. Russ Altman, Stanford University, and Dr, Branislav Vajdic, CEO of HeartBeam for encouragement and support for the work on this topic.